\documentclass[letterpaper,10pt,conference]{ieeeconf}

\IEEEoverridecommandlockouts                              
                                                         
\usepackage{graphicx} 
\usepackage{xcolor}
\usepackage{algorithm}
\usepackage{algorithmic}
\usepackage{amsmath}
\usepackage{amssymb}
\usepackage{cite}
\usepackage{booktabs}
\usepackage{stfloats}
\usepackage{makecell}
\usepackage{subcaption}
\usepackage[font=small]{caption}
\usepackage{tikz}
\usetikzlibrary{arrows.meta, positioning}
\title{\LARGE \bf ROSplane 2.0: A Fixed-Wing Autopilot for Research}
\author{ Ian Reid$^\dagger$, Joseph Ritchie$^\dagger$, Jacob Moore$^\dagger$, Brandon Sutherland$^\dagger$, Gabe Snow$^\dagger$, Phillip Tokumaru$^\ddagger$, Tim McLain$^\dagger$
\thanks{$^{\dagger}$Brigham Young University}
\thanks{$^{\ddagger}$AeroVironment Inc.}
}
\date{\today}

\begin{document}

\maketitle

\begin{abstract}
Unmanned aerial vehicle (UAV) research requires the integration of cutting-edge technology into existing autopilot frameworks.
This process can be arduous, requiring extensive resources, time, and detailed knowledge of the existing system.
ROSplane is a lean, open-source fixed-wing autonomy stack built by researchers for researchers. 
It is designed to accelerate research by providing clearly defined interfaces with an easily modifiable framework.
Built around ROS 2, ROSplane allows for rapid integration of low or high-level control, path planning, or estimation algorithms.
A focus on lean, easily-understood code and extensive documentation lowers the barrier to entry for researchers.
Recent developments to ROSplane improve its capacity to accelerate UAV research, including the transition from ROS 1 to ROS 2, enhanced estimation and control algorithms, increased modularity, and an improved aerodynamic modeling pipeline.
This aerodynamic modeling pipeline significantly reduces the effort of transitioning from simulation to real-world testing without requiring costly system identification or computational fluid dynamics tools.
ROSplane's architecture reduces the effort required to integrate new research tools and methods, expediting hardware experimentation.
\end{abstract}

\section{Introduction} \label{Introduction}

Unmanned aerial vehicles (UAVs) have gained significant popularity due to applications such as package delivery, photography, surveillance, or advanced air mobility (AAM).
Research targeting UAVs often requires ready access to the inner workings of many portions of an autonomy software stack, including estimation, path planning, and high/low-level control.
This access is important in all stages of research and development, both in simulation and hardware flight tests.

Researchers often face significant integration challenges when it comes to conducting realistic simulations and real-world flight tests.
Research software can require extensive rewriting or refactoring to be compatible with realistic simulations or hardware platforms.
This increases the difficulty of conducting real-world experiments, which are essential steps in validating and proving out new research algorithms and methods.

ROSflight\cite{rosflight2025} is a lean, open-source\footnote{https://github.com/rosflight} autopilot designed to mitigate these challenges and reduce the barrier to entry for UAV research.
It accomplishes this by allowing the same software that runs in simulation to also control the physical vehicle with no changes.
Since ROSflight is built on the Robot Operating System (ROS 2)\cite{ros2_2022}, it offers superior modularity and customizability.

ROSplane\cite{rosplane2017} is an open-source\footnote{https://github.com/rosflight/rosplane} autonomy stack for fixed-wing UAVs designed to work with ROSflight.
A lean feature set means ROSplane offers not only basic functionality, but also enables better understanding for quick and seamless integration of external codebases.
Extensive documentation on the algorithms used in ROSplane is available, making it a valuable resource for educational use and research\cite{uavbook}, \cite{rosplane2017}.

While ROSplane has received detailed attention in the past\cite{rosplane2017}, recent advances have significantly improved ROSplane's use in advanced UAV research.
These improvements reduce the barriers to practical UAV research by enhancing the usability, modularity, and extensibility of ROSplane, especially for hardware experiments.
The contributions of this work are to describe the advancements of ROSplane including 
\begin{itemize}
    \item the transition from ROS 1 to ROS 2,
    \item updated algorithms and modularity for control and state estimation,
    \item an improved aerodynamic modeling pipeline that significantly reduces the simulated-to-real experiment transition effort.
\end{itemize}

The rest of this work is organized as follows.
Section \ref{sec:related-work} discusses other available autopilots and related work.
Section~\ref{sec:system-arch} describes the system architecture.
The improved aerodynamic modeling pipeline is described in Section \ref{sec:sim2real}.
Algorithm improvements are discussed in Section \ref{sec:algorithm}.
A tutorial on using ROSplane is provided in \ref{sec:tutorial}.
Hardware test results and comparison to simulation are shown in Section \ref{sec:results}, and then concluding remarks are offered.

\section{Related Work}\label{sec:related-work}
Many excellent autopilots and flight control software are available for hobby, commercial, and research users alike.
Current open-source autopilots like PX4\cite{px4} and ArduPilot~\cite{ardupilot} offer excellent \textit{plug-and-play} capability.
They have the advantage of years of development, large communities, and full-featured autonomy stacks.
Although they offer distinct advantages, these autopilots have large code bases not designed for easy modification or integration, and can require a researcher to gain familiarity with these large code bases to implement given research.
It can also lead to a \textit{black box} environment by obfuscating core autopilot features, making integration of research code difficult and time consuming.
Furthermore, code for these autopilots runs primarily on embedded microcontrollers, which makes debugging and active development more difficult and microcontroller-dependent.

Even excellent autopilots like Paparazzi that are designed for researchers have limited ability to directly change underlying code.
Instead, the project opts for generation of new autopilot modes through XML files.
Paparazzi also lacks critical integration that may be necessary for advanced simulation \cite{autopilot_survey}.

ROSplane 2.0 offers an autonomy stack that is designed for easy integration with fixed-wing UAV research of any kind.
Its lean feature set allows the code to be easily understood, which lowers the integration time and effort for researchers, thus also lowering the barrier to entry and the learning curve.
ROSplane also moves the entirety of the autopilot stack from the embedded microcontroller-based flight control unit (FCU) to the Linux-based companion computer.
This facilitates research code development and means that no change of firmware-level code is required, the code is easily modifiable, and transition from simulation to flight is seamless.
Pairing this with powerful tools like Docker \cite{docker} simplifies testing and development transitions from vehicle to vehicle.
The ROSplane architecture is described more fully in Section \ref{sec:system-arch}.

ROS 2 has been extensively used for research and commercial use alike\cite{ros2_2022}.
Existing autopilots like PX4 have made significant efforts to support ROS 2 software, and integration with ROS 2 software and PX4 is constantly improving \cite{autopilot_survey}.
ROSplane takes ROS 2 support a step further by structuring all autopilot features as ROS 2 nodes, including controllers, estimators, and path planners.
This tight integration with ROS 2 significantly improves customizability and modularity.

Another challenge with UAV research is the lack of defined aerodynamic models for many UAVs.
This limits the ability of researchers to perform accurate testing in simulation.
Further, accurate aerodynamic analysis software is often expensive, which only exacerbates barriers to entry in UAV research.
Some recent works have explored using open source aerodynamic modeling and analysis tools for UAV system identification \cite{avl_testing}, \cite{Barrella2024}.
However, there are few clear documentation and tutorials that assist researchers through the whole process of designing the model, performing stability and control analyses, executing in simulation, and transitioning to hardware.
ROSplane 2.0 provides a simulation environment with easily modifiable aerodynamic parameters.
Additionally, ROSflight 2.0 provides an aerodynamic modeling pipeline with complete documentation that takes users through this process using two readily-available tools, XFlyer5 \cite{Xflr5} and OpenVSP \cite{OpenVSP}.

\section{System Architecture}\label{sec:system-arch}
ROSplane's system architecture is designed to accelerate deployment of advanced research by allowing researchers to more easily integrate research code, test in a simulation environment, and deploy on real hardware.

\subsection{Transition from ROS 1 to ROS 2}
ROSplane 2.0 has been updated to use ROS 2 instead of ROS 1.
Long-term support (LTS) versions of ROS 2 are primarily supported, including ROS 2 Humble on Ubuntu 22.04 and ROS 2 Jazzy on Ubuntu 24.04.
Other non-LTS version of ROS 2 are not officially supported.

ROS 2 offers significant advantages over ROS 1, including increased modularity and flexibility for integrating new features, sensors, and code \cite{ros2_2022}.
ROSplane 2.0 builds off the capabilities of ROS 2 by distributing the autonomy stack and the communication paths between independent nodes.
This modular structure decouples responsibilities in the code, preventing changes in one module from unintentionally propagating through the control stack, aiding researchers integrating new algorithms.
This ROS 2-defined structure also reduces the complexity of integrating novel sensors and external simulators by allowing each component to be added as an independent node.
This allows researchers to test new algorithms or hardware without needing to completely restructure logic or interfaces.

\subsection{Autopilot Structure}
\begin{figure}
    \centering
    \includegraphics[width=0.8\linewidth]{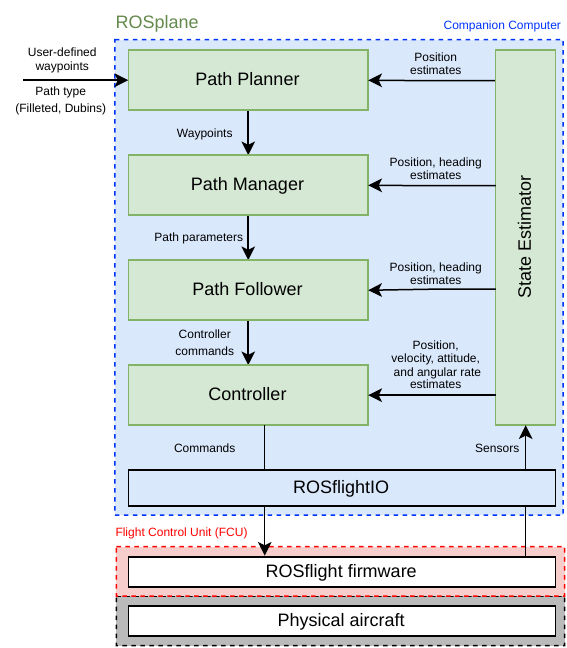}
    \caption{Cascaded architecture of ROSplane's modules.}
    \label{fig:system_architecture}
\end{figure}

The default implementation of ROSplane 2.0 follows the framework of \cite{uavbook}, as shown in Figure \ref{fig:system_architecture}.

This architecture gives ROSplane basic waypoint-following functionality out-of-the-box.
Due to the design philosophy of ROSplane, the waypoint-following functionality is neither complex, advanced, nor fully-featured.
Instead, the structure of ROSplane facilitates easy understanding with clearly delineated responsibilities assigned to each module and lean, well-understood implementations.
This makes it easier for users to identify which modules are necessary for their specific applications and which modules should be removed, replaced, or modified, thus increasing modularity and customizability while reducing time invested in deployment.

Figure \ref{fig:system_architecture} shows that ROSplane follows a cascaded architecture as the output of one module feeds into the next node, with the estimator providing state estimates to each.
The responsibility of the path planner module is to generate high-level waypoints that are consumed by the path manager.
In the default implementation of ROSplane, the path planner simply aggregates user-defined waypoints and publishes these waypoints to the path manager.
The path manager is responsible for managing which waypoints are active by determining when the UAV has achieved a waypoint and what type of path (straight line or orbit) is required to fly to the next waypoint.
The path manager also determines if the paths followed are Dubins paths, and calculates the correct path parameters.
The path manager then sends path definitions to the path follower node, whose responsibility is to follow these path commands by generating lower-level controller commands.
These lower-level commands are composed of desired airspeed, altitude, and course setpoints, which are used by the controller module to determine the desired control surface deflections and the throttle setpoint.
ROSplane's estimator module receives sensor information through the ROSflightIO node (a node responsible for managing communication between the companion computer and the FCU, as discussed in \cite{rosflight2025}).
It is responsible for constructing a state estimate, which is then sent to all other modules.
A more detailed description of each module is found in \cite{uavbook}.

Each of these modules is implemented as a ROS 2 node and all communication over the modules occurs over the ROS 2 network.

\section{Simulation to Real World Workflow}\label{sec:sim2real}

This section describes an aerodynamic modeling pipeline that significantly improves the simulation-to-real transition.
The aerodynamic model used is described in \cite{uavbook}.

Tuning a controller or attempting new research on an aircraft for the first time during a physical flight test can be dangerous because the aircraft's performance has not been validated. 
Unstable responses are possible and potentially catastrophic.
Tuning or testing new capabilities in simulation before flight is preferable because it avoids these risks and demonstrates real-world performance; however, an accurate aerodynamic model and simulation are necessary. 
Unfortunately, accurate aerodynamic models of UAVs are difficult to obtain with traditional methods, as system identification is time-consuming, expensive, and often requires extensive flight data augmented with significant wind tunnel testing.  
Rigorous computational tools (e.g. computational fluid dynamics) are also available to help with system identification, but they are often costly, slow, and require significant researcher effort.
These challenges make it difficult for researchers to test new autonomous aircraft.

The ROSplane simulation is capable of providing accurate controller tuning with a lower-fidelity aerodynamic model that can be obtained using open-source aircraft modeling software.
Documentation on the aerodynamic modeling pipeline and ROSplane integration steps are available on the project website\footnote{rosflight.org}.

Several open-source software tools for aerodynamic modeling and analysis are publicly available. 
Notable examples include XFLR5 \cite{Xflr5} and OpenVSP \cite{OpenVSP}, both parametric aircraft modeling and aerodynamic analysis programs.
Both tools enable users to define aircraft geometry and conduct a range of aerodynamic analyses to acquire aerodynamic parameters and stability and control derivatives \cite{aircraft_modeling}. 
The steps involved in the aerodynamic modeling pipeline are outlined in the following flowchart. 

\begin{figure}[htbp]
    \centering
    \includegraphics[width=\linewidth]{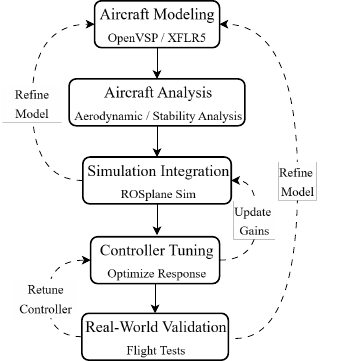}
    \caption{Aircraft integration pipeline with iterative feedback.}
    \label{fig:flowchart}
\end{figure}

When modeling and analyzing an aircraft, users should pay careful attention to ensure that accurate aircraft dimensions, flight conditions, and trim-state conditions are used. 
Once the aircraft model is complete, the aerodynamic parameters and stability and control derivatives can be obtained by performing an aerodynamic analysis and a stability analysis for each set of control surfaces. 
The aircraft model can then be integrated into the ROSplane simulation environment by inputting all of the aerodynamic coefficients into an aerodynamic parameters file. 
After loading the model in simulation, the researcher can then tune both the aircraft model and the flight controller. 
Both XFLR5 and OpenVSP have been shown to provide sufficiently accurate aircraft models for tuning flight controllers in close to trim-state conditions; however, some errors in the model can be present and will likely lead to inaccurate simulation performance \cite{aerodynamic_errors}.

Before tuning a flight controller, users should validate the accuracy of their aerodynamic model by manually flying the aircraft in the simulation. 
If the aircraft has unrealistic behavior, then the aircraft model may require multiple cycles of improvement. 
Users can often improve their aircraft model by eliminating errors in meshing, adding additional details (contours and smaller features), and increasing the precision of dimensions while comparing against experimental results. 
After several cycles of iteration, both XFLR5 and OpenVSP models should provide sufficient fidelity to support effective pre-flight controller tuning that is safer than tuning the controller blindly on a maiden flight. 

To tune the flight controller, users can set the flight controller to control the aircraft and observe the controller responses to a variety of waypoint, trajectory, rate, and position commands. 
Users can then modify any controller gains during flight through ROS terminal commands, or the RQT GUI, until the controller behavior is sufficient.
Once the flight controller is tuned to provide appropriate responses in simulation, it is well-prepared to safely provide control during a physical flight. 

This approach reduces tuning risk, accelerates controller development, and promotes reproducible UAV research without requiring expensive wind-tunnel testing or high-end computational resources.

\section{Modularity and Algorithmic Improvements}\label{sec:algorithm}

ROSplane 2.0 has improved the modularity of the system by making useful functionality more accessible to the end user.
The state estimator and controller have been restructured with clearly defined interfaces and internal structure to enable safe and effective modifications without a complete overhaul.
Both the the controller and estimator have also had substantial algorithmic changes that improve performance.
The algorithmic changes follow the most recent changes reflected in \cite{uavbook}.
They represent improved accuracy and robustness in the attitude and velocity state estimates, along with improved path following.

 \subsection{State Estimation}
Previously, ROSplane used a pedagogically-motivated two-stage estimation scheme based on the implementation in \cite{uavbook}.
It separated attitude estimation and positional states.
This made it easier for students to implement \cite{rosplane2017}; however, the estimator was less accurate than desired.
Additionally, while the implementation in \cite{uavbook} required a measurement of heading (e.g. digital compass or magnetometer), the previous ROSplane estimator did not implement one, resulting in poor wind estimates.
In ROSplane 2.0, the estimator has been replaced with a full-state extended Kalman filter (EKF).
This full-state EKF has an expanded number of estimated states, including heading, gyroscope biases and lateral velocity.
The estimator also takes full advantage of GNSS velocity measurements and measurements from the magnetometer.
The updated estimator shares its structure and derivation with the one presented in ROScopter with the addition of two measurement updates \cite{roscopter}.
The measurement updates included in ROSplane are the updates for the differential pressure and the zero side-slip angle assumption that drive wind estimation.

The sensor reading $z_{\text{diff}}$ measures the differential pressure from the pitot tube.
The measurement model for the differential pressure update $h_{\text{diff}}(x)$, derived from the state $\boldsymbol{x}$, is given by
\begin{equation}
    h_{\text{diff}}(\boldsymbol{x)} = \frac{1}{2} \rho \big(\boldsymbol{v} - R^\top(\boldsymbol{\theta})\boldsymbol{w}\big)^\top \big(\boldsymbol{v} - R^\top(\boldsymbol{\theta})\boldsymbol{w}\big),
\end{equation}
where $\boldsymbol{v}$ is the estimated velocities in the body frame, $R(\boldsymbol{\theta})$ is the rotation matrix that rotates values in the body frame into the inertial frame and $\boldsymbol{w}$ is the estimated wind in the inertial frame. 

To enforce the nominal dynamics of the aircraft that the side-slip angle $\beta$ is zero, the estimator uses a pseudo-measurement update.
This assumes that all velocity along the body y-axis of the airplane is caused by wind.
The pseudo-measurement $z_{\beta}$ is taken to be zero and the measurement model $h_{\beta}(\boldsymbol{x})$ is given as,

\begin{equation}\label{eq:sideslip}
    h_\beta(\boldsymbol{x}) = \begin{bmatrix} 0&1&0 \end{bmatrix}(\boldsymbol{v} - R(\boldsymbol{\theta})^\top\boldsymbol{w}).
\end{equation}

Eq. \eqref{eq:sideslip} finds the estimated side-slip velocity while taking away the effects of wind, which should nominally be zero.
These measurement updates are incorporated as in \cite{roscopter} and drive the wind estimation portion of the ROSplane estimator.

The EKF also has modularity improvements.
Beyond being completely replaceable with the simple output interface, the EKF is designed to easily accommodate new measurement updates, allowing the researcher to easily integrate new sensors.
The framework for adding new sensors can be found on the project website.

\subsection{Control}
The controller architecture for ROSplane 2.0 has also been improved.
Previously, the altitude was controlled using a state machine to command a climb rate and descent rate if outside of a band around the cruising altitude.
The altitude controller has been unified into a successive loop closure controller, improving altitude response. 
ROSplane 2.0 also integrates the improvements in orbit following.
A feedforward term is added to commanded roll angles to allow for faster and tighter convergence on orbit paths.
In addition, ROSplane 2.0 makes available a total energy controller for use.
Section \ref{sec:results} shows the performance of the successive loop closure controller in real flight conditions.

\begin{figure}[htbp]
    \centering
    \includegraphics[width=\linewidth]{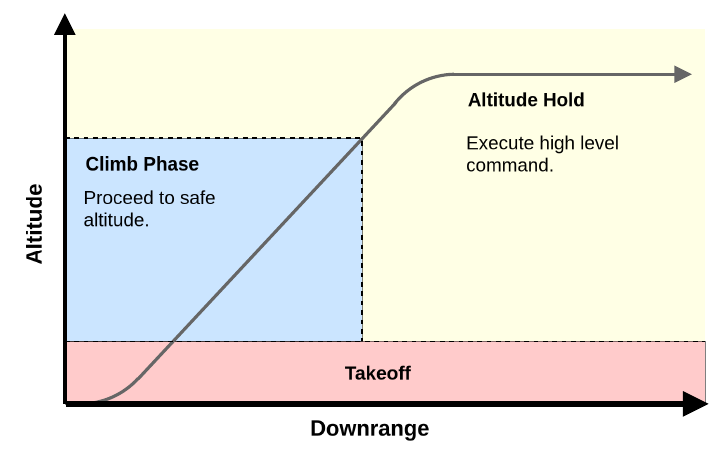}
    \caption{The state machine for different phases of flight.}
    \label{fig:control_sm}
\end{figure}

The controller's modularity improvements allow for safe integration of new control schemes.
This includes utilizing a state machine to ensure that full autonomous control is only taken under safe conditions.
ROSplane 2.0 has a simple state machine to allow for the controller to have different control schemes based on the altitude and phase of flight.
These are summarized in Figure \ref{fig:control_sm}.
Integration with these phases of flight and the modular nature of the controller maximize flexibility for the researcher to integrate research by allowing them to take as much or as little control as they need.
A researcher can allow ROSplane's native behavior to dominate in climb then transfer control to a new in development algorithm only when cruise has been achieved.

\section{Tutorial}\label{sec:tutorial}
This section contains a brief tutorial on how users can expect to use ROSplane to aid their research.
This is not a comprehensive tutorial, meaning that the ideas listed here serve as a starting point.
For a more detailed description please see the project website.
We will first describe ROSplane's default functionality, and will then give examples of how to customize ROSplane to fit individual needs.
Finally, we will discuss the intended workflow for researchers to use ROSplane in their research.

\subsection{Default High-level Functionality}
This section describes ROSplane's basic functionality that can be used out of the box.
By default, ROSplane enables an aircraft to fly waypoint missions.
These waypoints are simply 3D coordinates, and can be specified in latitude, longitude, and altitude (LLA), or in meters north, east, and down (NED) from the initial position of the aircraft.
These waypoint types can be mixed and matched---for example, one waypoint can be specified in LLA format and the next can be specified in NED format.
Waypoints are loaded using service calls to ROSplane's path planner module, as shown in Figure \ref{fig:system_architecture}, and can be loaded one at a time, or all at once as specified in a YAML file.

ROSplane's default path planner module can be used by high-level planning or coordination algorithms.
For example, a high-level exploration module for search-and-rescue might specify waypoints that an aircraft needs to fly, or a multi-agent coordination algorithm might specify a rendezvous point as a single waypoint.
These algorithms may not need more complex path-following than the straight lines and orbits in ROSplane.
In this case, the default ROSplane functionality could effectively sit underneath a researcher's high-level algorithm.

\subsection{Customizing ROSplane}
ROSplane's basic waypoint-following capability will not satisfy all researchers' needs.
Because of this, ROSplane has been designed so that each module has a single, specific responsibility, as shown in Figure \ref{fig:system_architecture}.
This separation of responsibility makes it easy to determine which modules can be removed or modified and which modules should be kept for a given application.
Additionally, ROSplane exploits the modular nature of ROS 2 to allow these modules to be easily modified without affecting the rest of the autonomy stack.

Each module in ROSplane uses an interface class to define each module's ROS 2 interfaces, as well as the required functions that a derived node must implement.
If a given derived class successfully inherits from the base interface class, it will integrate seamlessly with the rest of ROSplane.
For example, the ROSplane path follower module uses a vector field approach from \cite{uavbook} to follow straight-line or orbit paths that are specified by the path manager module.
Changing the way ROSplane follows these straight-line and orbit paths can be done by simply creating a node that has the same ROS 2 interfaces as the original path follower.
This can be done by either inheriting from the base class in the node, or reimplementing the ROS 2 interfaces if inheriting is not possible (e.g. if using Python instead of C++).

Many times, users will desire to customize ROSplane in a way that crosses the boundaries of ROSplane's default structure.
For example, a researcher studying different spline-following methods would need to modify ROSplane so that it follows splines instead of straight lines and orbits.
Thus, both the path manager and path follower modules would need to be replaced.
The linear flow of information through ROSplane makes this easy---as long as the replacement module has the same input ROS 2 interfaces (subscribers and service servers) as the path manager and the same output ROS 2 interfaces (publishers and service clients) as the path follower, the rest of ROSplane will seamlessly work with the new application code.

ROSplane has been designed to be as flexible as possible so that modules can be replaced, removed, or modified as needed for a particular application.

\subsection{Intended Workflow for ROSplane}
This section describes the intended workflow for users interested in ROSplane.
The first step is for a researcher to develop their application-specific code and integrate it into the ROSplane autonomy stack.
Then, the aircraft used in hardware flight tests should be modeled as described in Section \ref{sec:sim2real}.
If hardware tests will not be performed, then the default aerodynamic model can be used.
It is recommended that users then use the ROSflight simulation to test and further refine the application code.
The project website has detailed information on setting up the simulation environment.
Once the application code works well in simulation, hardware tests can be performed.
Because ROSflight and ROSplane run the exact same code in simulation and in hardware \cite{rosflight2025}, these hardware tests can be performed in the exact same manner as the simulation tests were performed.

\section{Results}\label{sec:results}

This section presents the results of simulation and hardware experiments demonstrating the basic functionality ROSplane offers to users out-of-the-box.
These results also demonstrate that ROSplane closes the simulation-to-real gap for research.

\subsection{Simulation}

\begin{figure}[htbp]
    \centering
    \includegraphics[width=0.9\linewidth]{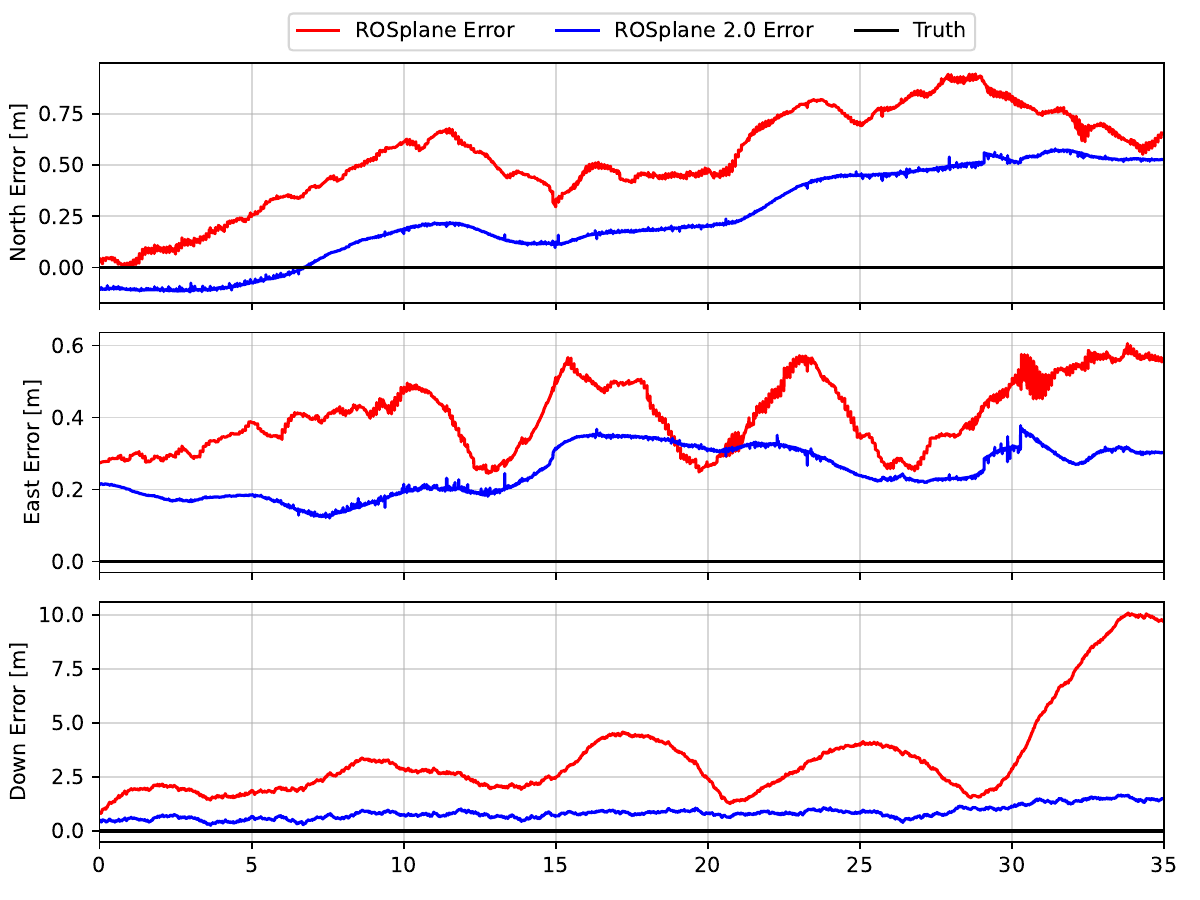}
    \includegraphics[width=0.9\linewidth]{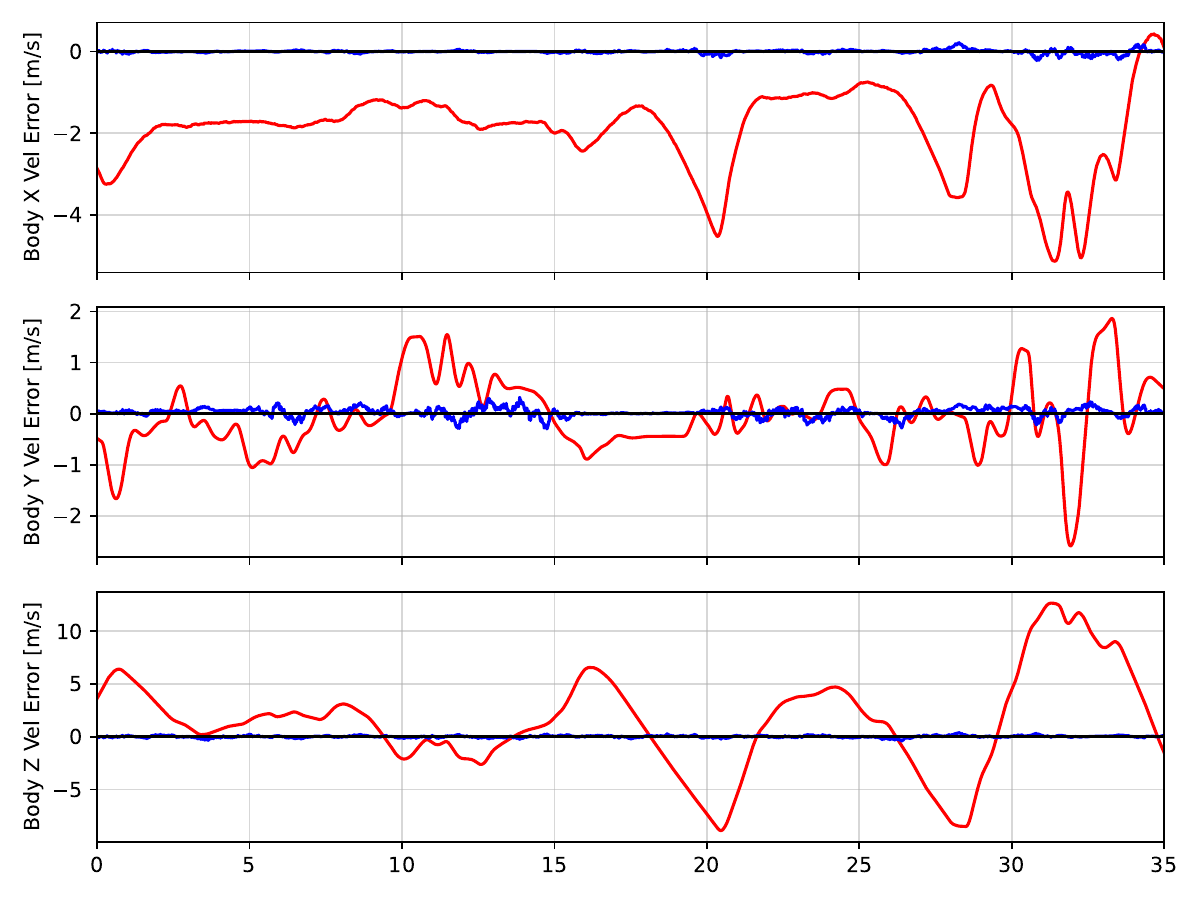}
    \includegraphics[width=0.9\linewidth]{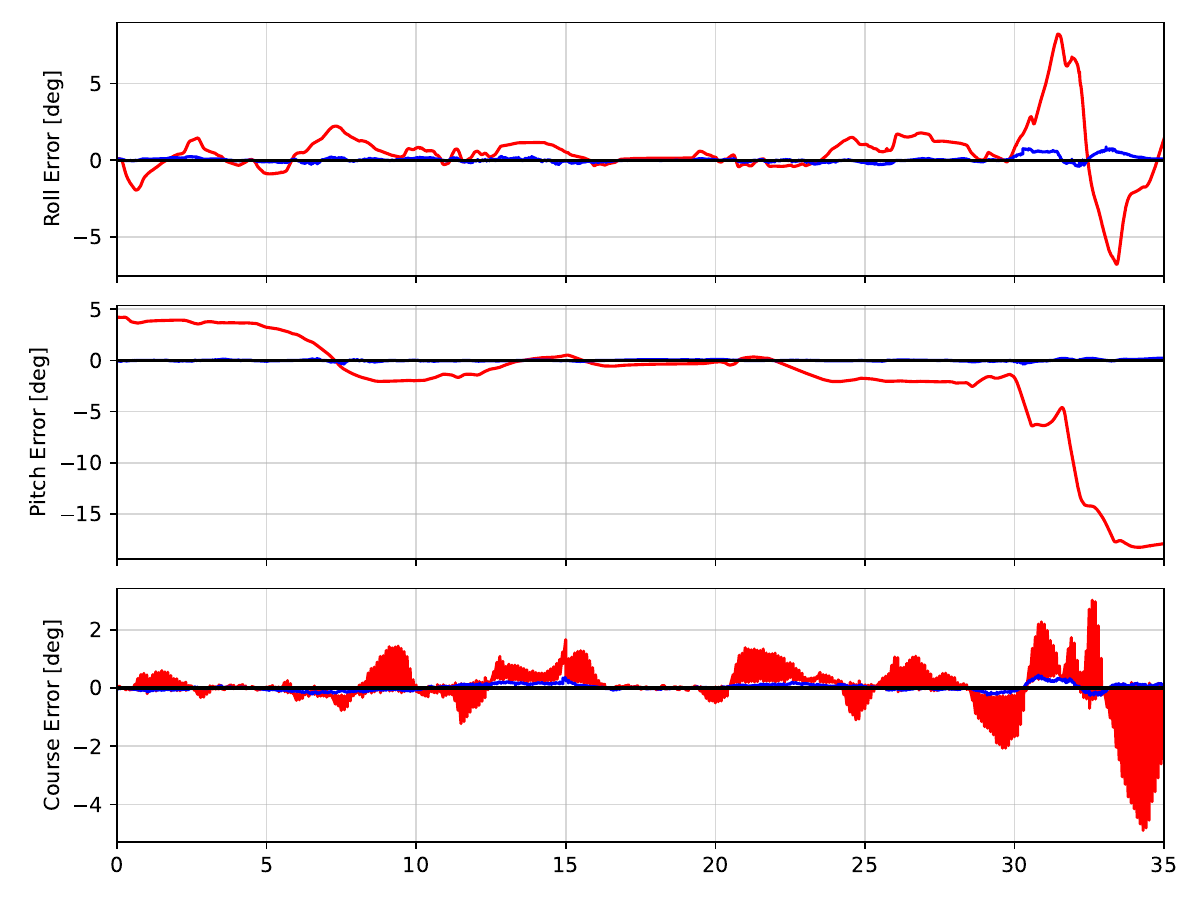}
    \includegraphics[width=0.9\linewidth]{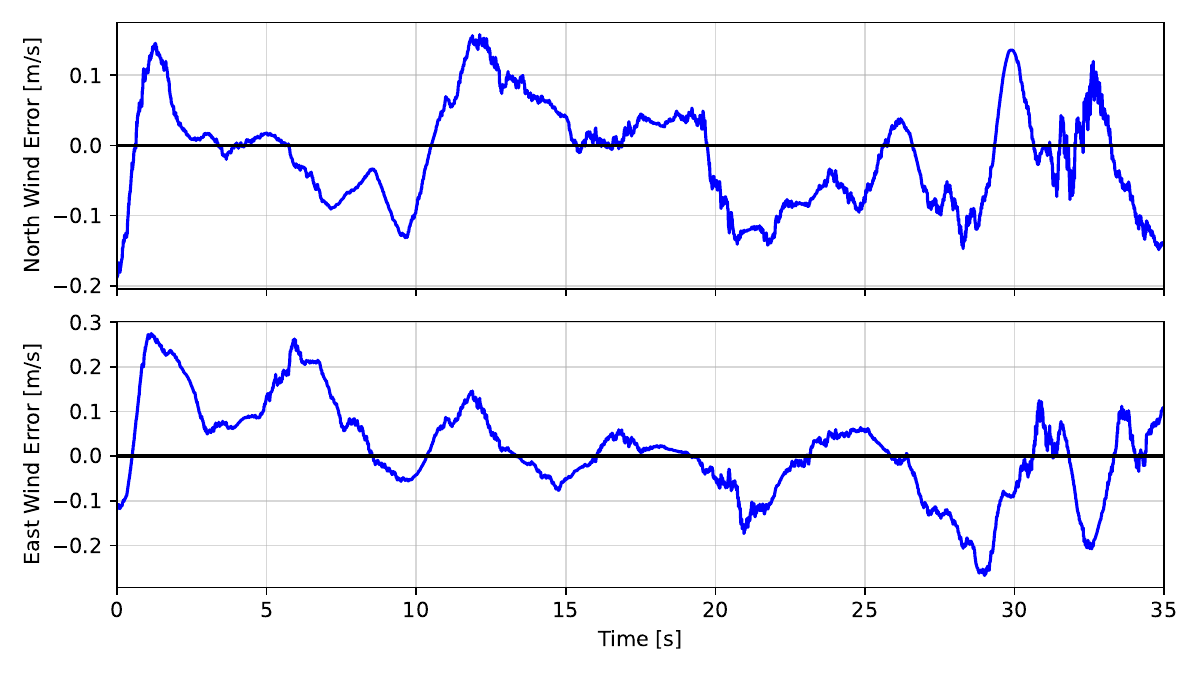}
    \caption{Comparison of ROSplane and ROSplane 2.0 estimated state errors during a manually controlled flight in simulation.}
    \label{fig:old_vs_new}
\end{figure}

The ROSplane 2.0 full-state estimator was first compared to the previous ROSplane estimator in a simulation environment.
An aircraft was flown manually in simulation and the sensor data was recorded and fed through both estimators.
Figure \ref{fig:old_vs_new} compares the errors in estimated state against ground truth.
Wind estimates are shown only for the ROSplane 2.0 estimator due to poor performance in the previous version which did not implement a magnetometer.
Table \ref{tab:estimator-rmse} shows RMS error between all estimated states for both estimators.
The full-state EKF has significantly less tracking error in attitude, velocity and position, and demonstrates that it provides consistent and reliable estimation to support the flight experiments of researchers.
In the companion paper for ROScopter, the improved estimator is shown to perform well compared against the PX4 estimator \cite{roscopter}.

\subsection{Simulation to Real Flight}

The simulation capabilities of ROSplane and the aerodynamic modeling pipeline were tested according to the aircraft integration process outlined in Section \ref{sec:sim2real}.
The responses of the controller in simulation were then compared to those recorded during a physical flight, given the same trajectory commands and controller configuration. 

\begin{figure}[htbp]
    \centering
    \includegraphics[width=\linewidth]{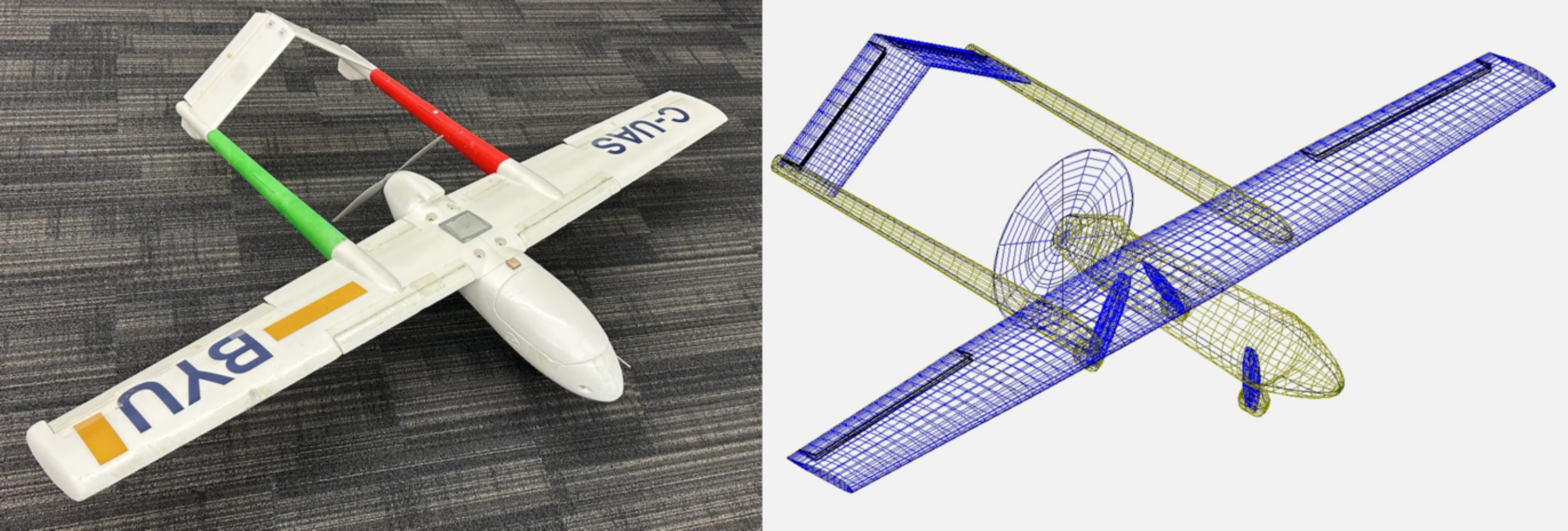}
    \caption{RMRC Anaconda UAV (left) with digital OpenVSP model used for aerodynamic analysis (right).}
    \label{fig: anaconda pic}
\end{figure}

\begin{figure}[htbp]
    \centering
    \includegraphics[width=\linewidth]{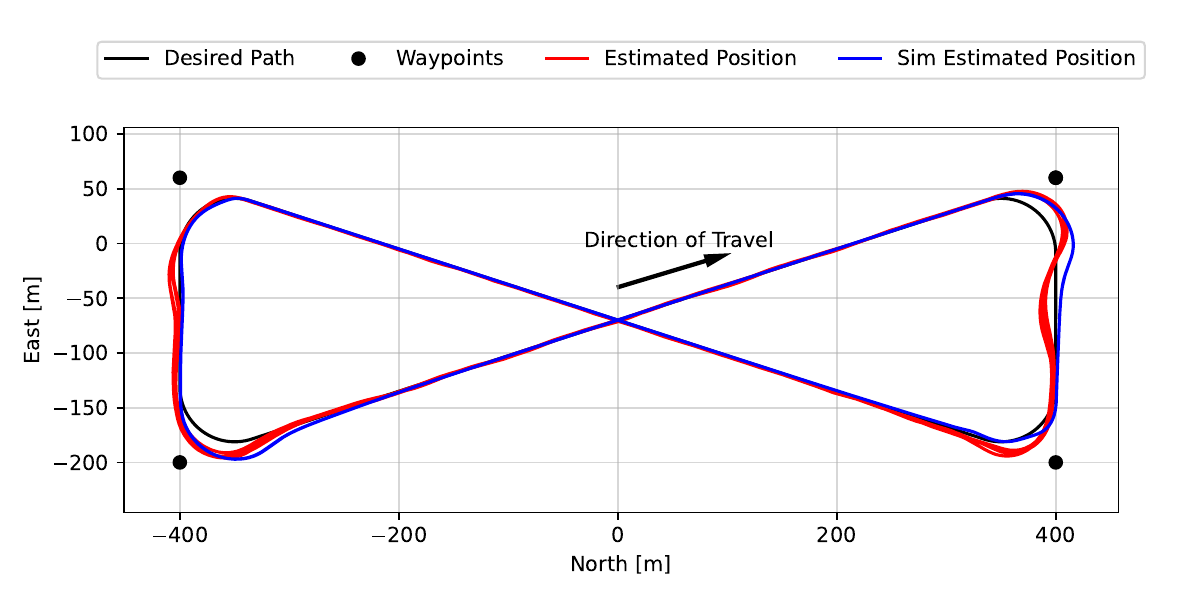}
    \caption{Comparison of a commanded flight path with OpenVSP VLM-derived simulation and several full flight paths of physical comparison results.} \label{fig:flight_path_comparison}
    \vspace{-15pt}
\end{figure}

A model of the RMRC Anaconda UAV shown in Figure \ref{fig: anaconda pic} was created in both XFLR5 and OpenVSP, and a variety of aerodynamic and stability analyses were performed to obtain the aerodynamic coefficients and stability and control derivatives for the aircraft.
XFLR5 and OpenVSP both produced aerodynamic models that were accurate enough for controller tuning, though each tool excelled in different areas. 
XFLR5’s built-in trimming capability made stability analyses straightforward and yielded more consistent stability derivatives. 
OpenVSP, by modeling the full aircraft geometry, provided more reliable drag estimates and control derivatives. 
For both the Xflr5 and OpenVSP models, several cycles of iteration, as outlined in Section \ref{sec:sim2real}, were required for the aircraft model to converge on realistic performance.
Figure \ref{fig:flight_path_comparison} shows the OpenVSP simulation results compared with several circuits from physical flight results.
The final performance of the simulation and physical flight was comparable.
Note that the same controller gains were used in both flights, meaning that there was no difference in ROSplane's controllers for this comparison.
The Xflr5 results were similar. 

These comparisons validate that new UAVs can be modeled and analyzed using open-source software, integrated into the ROSplane simulation, and used for controller tuning, mission planning, and research testing. 
Iterative testing and improvement of these models provided confidence that real-life system performance can be accurately predicted within a safe simulation environment.
This allows for high confidence in the success of research algorithms in real flight tests.

\subsection{Hardware}

\begin{figure}[htbp]
    \centering
    \includegraphics[width=\linewidth]{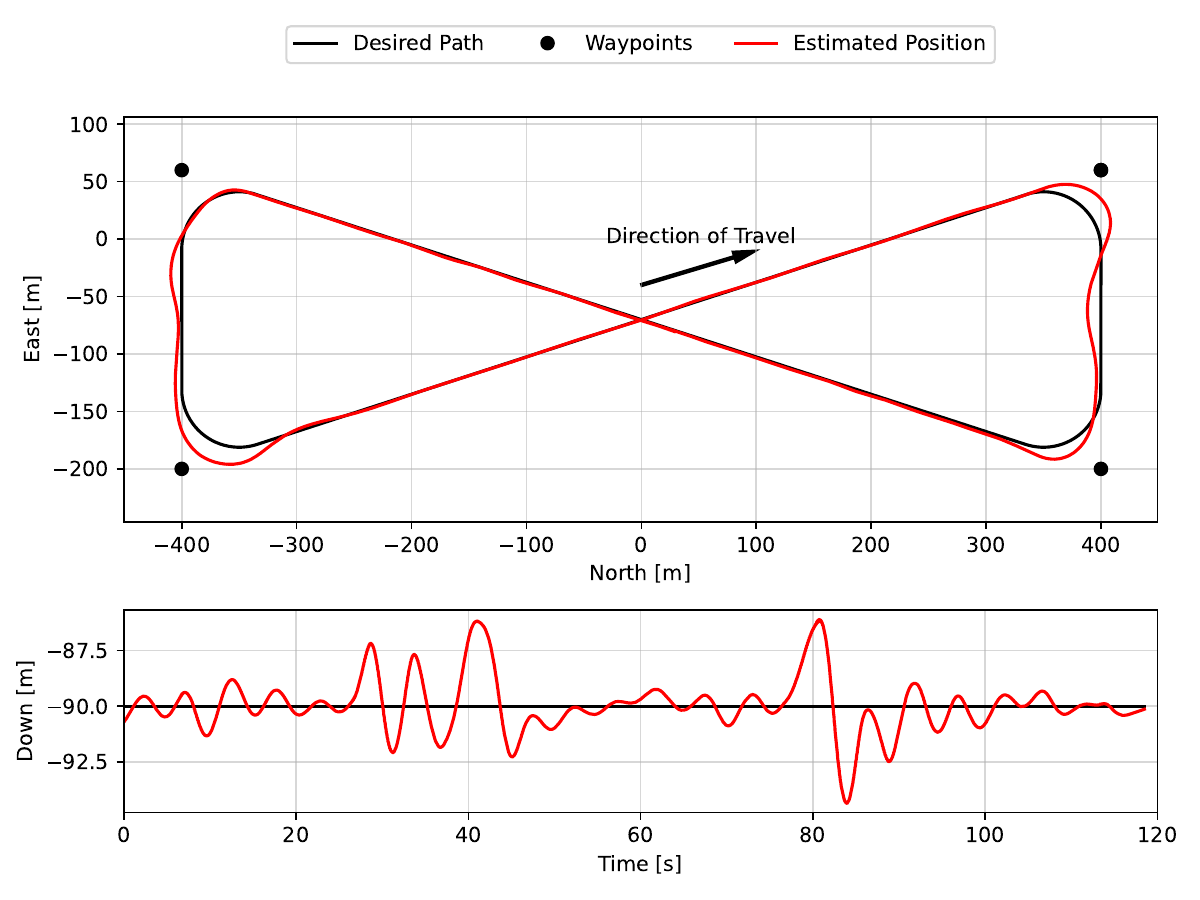}
    \caption{ROSplane estimated position alongside commanded path. Moderate winds from the south-east were present. ROSplane is able to successfully follow the waypoint mission.}
    \label{fig:hardware_loop}
\end{figure}

To validate the performance of ROSplane, a simple waypoint mission was flown on the RMRC Anaconda UAV shown in Figure \ref{fig: anaconda pic}.
ROSflight firmware was run on an FCU using the configuration 2 hardware \cite{rosflight2025}, which includes an onboard Jetson Orin as the companion computer.
In each test, the aircraft was directed to fly to a series of four waypoints in an hourglass formation on a continuous circuit, as shown in Figure \ref{fig:hardware_loop}.
RTK GPS data was collected during the flight to validate position estimates, but was not used as input to ROSplane's estimator.
The desired path and ROSplane's performance is shown in Figure \ref{fig:hardware_loop}, with an estimated 4-6 m/s wind from the south-east direction.
The path following algorithm overshoots the turns of the path because the commanded path assumes that the aircraft is only acceleration-limited and can achieve its maximum bank angle instantaneously.
The deviations from the commanded altitude in Figure \ref{fig:hardware_loop} correspond to periods during turn maneuvers, where a drop in altitude is expected due to the loss of lift when rolled.

\begin{figure}[htbp]
    \centering
    \includegraphics[width=\linewidth]{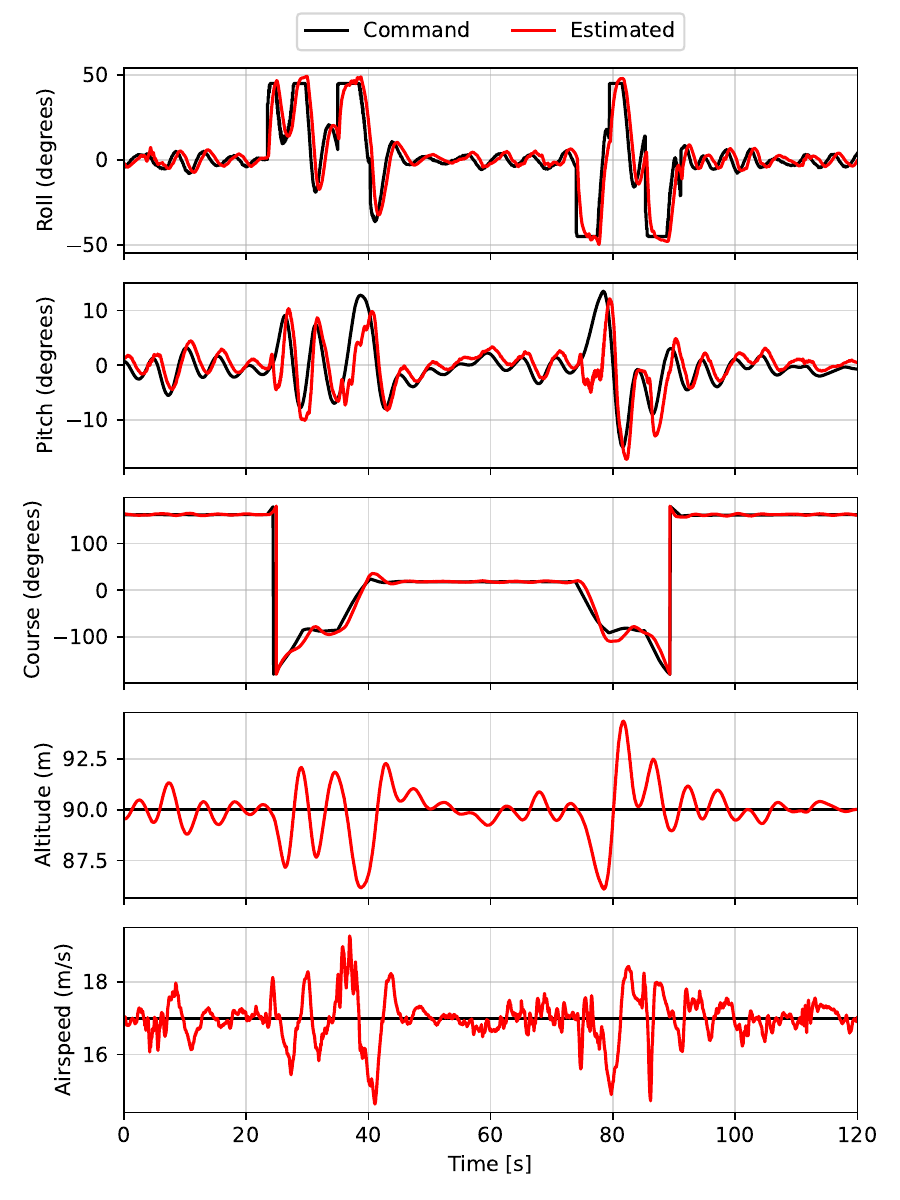}
    \caption{The controller response to given flight path input in the attitude angles, airspeed, and altitude. ROSplane effectively tracks the commanded setpoints.}
    \label{fig:controller_response}
\end{figure}

Figure \ref{fig:controller_response} shows the estimated states and the control setpoints for roll, pitch, course, altitude, and airspeed commands.
The aircraft was able to track the given commands effectively.
Times of poor response in the pitch loop are coincident with high roll commands, which is expected behavior for pitch-based altitude control.
It is important to note that the oscillations seen in the roll response have a frequency on the order of five seconds and are in response to disturbances in the course from the commanded trajectory.

\begin{figure}
    \centering
    \includegraphics[width=\columnwidth]{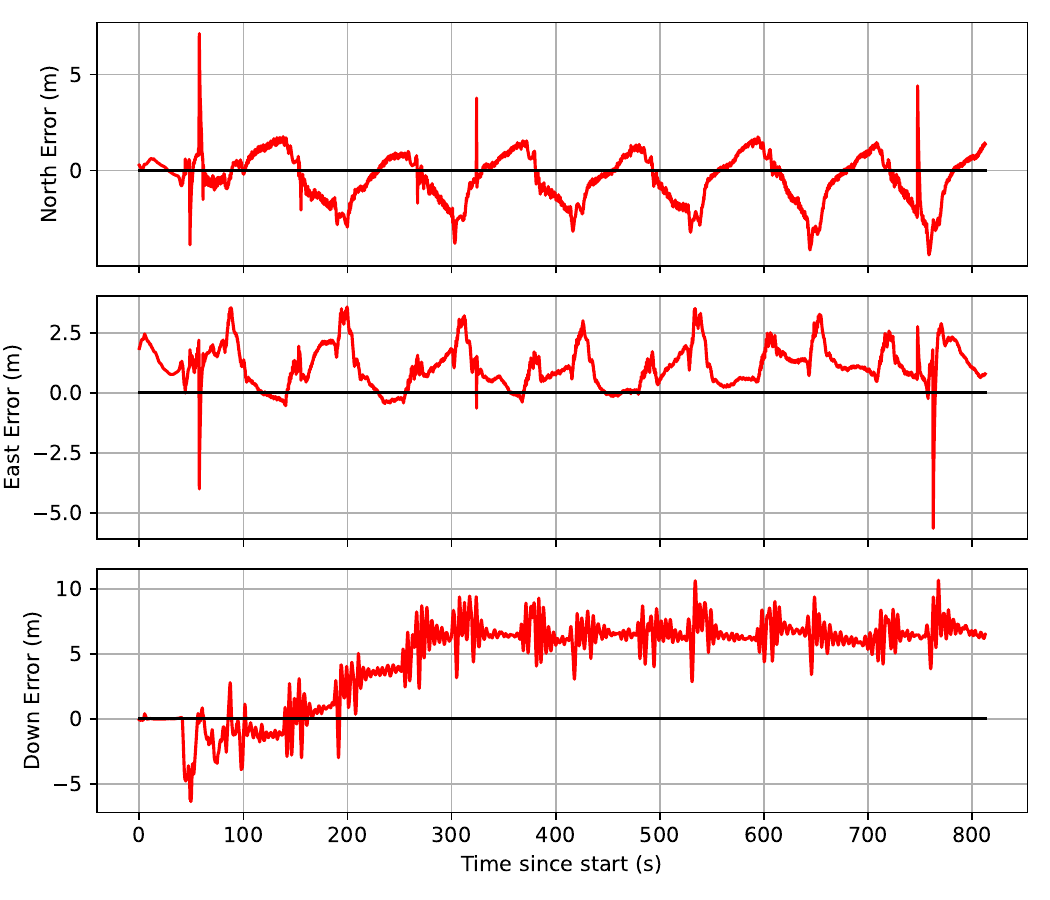}
    \caption{Error between the estimated position state and the RTK GPS data during portions of the flight with RTK GPS lock. RTK GPS was not used as an input to the estimator.}
    \label{fig:rtk_comparison}
    \vspace{-15pt}
\end{figure}

Table \ref{tab:estimator-position-rmse-vs-rtk} lists the RMS error between ROSplane's estimated positional states and RTK GPS, showing that ROSplane effectively estimated the vehicle's position.
Figure \ref{fig:rtk_comparison}
Ground truth for attitude and airspeed was not available.
Altitude error is significantly higher due to both barometric effects of wind speed on the cabin pressure and reduced vertical GPS accuracy.
Table \ref{tab:trajectory-rmse} shows RMS error of ROSplane's estimated state and the commanded trajectory, showing that ROSplane successfully tracked the commanded setpoints.
ROSplane is able to achieve consistent and reliably effective results in estimation and waypoint following.
These results show that ROSplane has a reliable basic feature set to be an effective platform for research.
Researchers can implement desired algorithms easily to evaluate different control, estimation, trajectory planning and following algorithms.
Full discussion and analysis of the controller and rationale for its performance can be found in \cite{uavbook}.

\begin{table}
    \centering
    \begin{tabular}{c|cccc}
             & North (m) & East (m) & Down (m) & Total (m) \\
        \toprule
        RMSE & 1.32 & 1.41 & 5.64 & 5.96 \\
        \bottomrule
    \end{tabular}
    \caption{Position RMS errors between ROSplane's estimated state and RTK GPS.}
    \label{tab:estimator-position-rmse-vs-rtk}
\end{table}

\begin{table*}
    \centering
    \begin{tabular}{c|ccccccc}
         & North (m) & East (m) & Altitude (m) & Roll (deg) & Pitch (deg) & Course (deg) & Airspeed (m/s) \\
         \toprule
         RMSE & 3.05 & 3.29 & 1.19 & 7.55 & 3.04 & 8.51 & 0.58 \\
         \bottomrule
    \end{tabular}
    \caption{RMS errors between the commanded trajectory and the estimated state during four loops of the waypoint mission flown in hardware and shown in Figure \ref{fig:hardware_loop}.}
    \label{tab:trajectory-rmse}
\end{table*}

\begin{table*}
    \centering
    \begin{tabular}{c|cccccccccc}
         Estimator & North (m) & East (m) & Altitude (m) & Vel-X (m/s) & Vel-Y (m/s) & Vel-Z (m/s)  & Roll (deg) & Pitch (deg) & Course (deg) \\
         \toprule
         ROSplane & 0.586 & 0.418 & 3.974 & 2.17 & 0.684 & 4.820 & 1.814 & 5.409 & 0.702 \\
         ROSplane 2.0 & 0.325 & 0.258 & 0.894 & 0.045 & 0.089 & 0.105 & 0.189 & 0.074 & 0.112 \\
         \bottomrule
    \end{tabular}
    \caption{RMS errors between estimated and simulated ground truth. The ROSplane 2.0 estimator significantly reduces RMS errors in estimated state.}
    \label{tab:estimator-rmse}
\end{table*}

\section{Conclusion}\label{sec:conclusion}
ROSplane's structure, capabilities, and modularity provide researchers with more flexibility and lower the barriers to entry to perform UAV autonomy research.
The improved code structure and architecture provide greater customizability and opportunities for integrating external hardware and code. 
ROSplane 2.0 architecture allows researchers to quickly and safely tune flight controllers and test algorithms for new fixed-wing UAV research in simulation to prepare for physical flight tests. 
This functionality of ROSplane 2.0 was demonstrated by modeling a UAV aircraft using open-source aircraft modeling and analysis software, integrating this aircraft model into the ROSplane simulation, and validating that the closed-loop controller performance of the simulation and physical flight are comparable. 
This enables researchers to perform accurate controller tuning and algorithm testing in simulation in the actual flight control software before flight, thereby reducing risk and iteration time.  
The universal applicability of the ROS 2 structure, paired with proven functionality and an effective simulation platform, makes ROSplane a powerful tool for researchers valuing customizability and direct control in cutting-edge autonomous research applications. 

\bibliographystyle{IEEEtran}
\bibliography{bibi}

\end{document}